\documentclass[letterpaper, 10 pt, conference]{ieeeconf}  % Comment this line out if you need a4paper

\IEEEoverridecommandlockouts                              % This command is only needed if 
\usepackage[T1]{fontenc}
\usepackage{graphics} % for pdf, bitmapped graphics files

\graphicspath{ {./Figures/} }
\usepackage{xcolor}

\usepackage{epsfig} % for postscript graphics files
\usepackage{amsmath} % assumes amsmath package installed
\usepackage{amssymb}  % assumes amsmath package installed

\newtheorem{assumption}{Assumption} % same for example numbers
\newtheorem{problem}{Problem} % same for example numbers

\usepackage{url}
\usepackage{hyperref}

\usepackage{cite}

\title{\LARGE \bf
CBF-Based Motion Planning for Socially Responsible Robot Navigation Guaranteeing STL Specification*
}

\author{Andrea Ruo, Lorenzo Sabattini and Valeria Villani% <-this % stops a space
\thanks{*This work was supported by Horizon Europe program under the Grant Agreement 101070351 (SERMAS).}% <-this % stops a space
\thanks{Department of Sciences and Methods for Engineering (DISMI), University of Modena and Reggio Emilia, Italy {\tt\small \{name.surname\}@unimore.it}}}

\begin{document}

\maketitle
\thispagestyle{empty}
\pagestyle{empty}

% INFO:
% Paper Length: 6 pages (8 pages for review purposes only)
% For the purpose of review only, manuscripts up to eight (8) pages in the standard two-column ECC conference format will be considered. If accepted, the final manuscripts are limited to a length of six (6) pages. Accepted papers exceeding the normal length may be included in the proceedings upon payment of over-length page charges for each page in excess of six pages, up to a maximum of two additional pages.
% Paper Size: US letter
% File Format: PDF
% Maximum File Size: 3MB

%%%%%%%%%%%%%%%%%%%%%%%%%%%%%%%%%%%%%%%%%%%%%%%%%%%%%%%%%%%%%%%%%%%%%%%%%%%%%%%%
\begin{abstract}
In the field of control engineering, the connection between Signal Temporal Logic (STL) and time-varying Control Barrier Functions (CBF) has attracted considerable attention. CBFs have demonstrated notable success in ensuring the safety of critical applications by imposing constraints on system states, while STL allows for precisely specifying spatio-temporal constraints on the behavior of robotic systems. Leveraging these methodologies, this paper addresses the safety-critical navigation problem, in Socially Responsible Navigation (SRN) context, presenting a CBF-based STL motion planning methodology. This methodology enables task completion at any time within a specified time interval considering a dynamic system subject to velocity constraints. The proposed approach involves real-time computation of a smooth CBF, with the computation of a dynamically adjusted parameter based on the available path space and the maximum allowable velocity. A simulation study is conducted to validate the methodology, ensuring safety in the presence of static and dynamic obstacles and demonstrating its compliance with spatio-temporal constraints under non-linear velocity constraints. 
 
\end{abstract}

%%%%%%%%%%%%%%%%%%%%%%%%%%%%%%%%%%%%%%%%%%%%%%%%%%%%%%%%%%%%%%%%%%%%%%%%%%%%%%%%
\section{INTRODUCTION}
%1 pg
\label{sec:introduction}
In recent years, several service robots have been developed for various practical applications, defining a novel approach to navigation called Socially Responsible Navigation (SRN). Notably, reception and robotic guidance have emerged as particularly popular services, where robots are gradually replacing human personnel in assisting customers. In these scenarios, a mobile robot autonomously navigates the environment to guide a person to a specific location, facing the challenge of planning and completing a collision-free path through obstacles in the environment \cite{LidarSlam}. 

Compared to robot navigation in non-social environments, such as underwater or warehouse environments, SRN takes into account both non-social obstacles and social agents, i.e., people, considering their comfort and social interactions \cite{onlineRobotNavigation}. In this application context, it is crucial to include safety-related constraints such as obstacle avoidance, velocity limits, and speed reduction when the robot is close to people. Additionally, space-time constraints might be relevant to ensure that the robot can efficiently and safely manage activities, especially in dynamic and crowded social environments shared with people. Temporal constraints can take various forms, such as time limits to complete a specific task, time intervals to complete a sequence of tasks, or priorities assigned to different activities based on their importance. These constraints may be imposed by environmental requirements or user requests.
Expressions such as ``\textit{the robot must reach the goal pose within 10 seconds}'' or ``\textit{the robot must remain within a specified area for 5 seconds}'' % or ``\textit{the robot must maintain a certain distance from the user}'' 
can be used to express such constraints. 
Temporal logics, like Signal Temporal Logic (STL) \cite{maler2004monitoring}, enable the specification of such spatio-temporal constraints, enhancing the expressiveness of Boolean logic through the temporal dimension. While STL has its roots in the field of formal verification in computer science, it is becoming increasingly popular as a well-established and systematic method for formulating spatio-temporal tasks in the field of control \cite{wiltz2022handling}.

A significant portion of the available control approaches for spatio-temporal tasks, as referenced in \cite{belta2007symbolic,9147796}, relies on automata theory, which can often be computationally intensive due to state discretization. As such, potential field-based methods can serve as a computationally efficient alternative for certain classes of spatio-temporal constraints.
In this regard, CBFs have recently garnered significant interest for safety-critical applications. By establishing a forward-invariant safe set through barrier functions and solving for control input using quadratic programming, CBFs ensure that the system remains within the safe set. CBFs provide a highly effective tool for designing controllers that are safe and computationally efficient \cite{liang2023control}. While early approaches to CBFs consider systems with a relative degree of one \cite{wieland2007constructive}, works in \cite{wiltz2022handling} and \cite{nguyen2016exponential} address systems with higher relative degrees. 

As a result, the connection between the semantics of an STL task and time-varying CBFs allows systems to be formally controlled while adhering to spatio-temporal constraints and ensuring safety. Several applications have demonstrated the potential of this combination, paving the way for innovative and advanced solutions. An innovative approach to integrate STL and CBF was presented in \cite{zehfroosh2022non}. 
In this work, the authors proposed a method that uses a navigation function as the foundation for constructing a CBF and combines barrier functions to encode Boolean operations among the predicates. In a different context, \cite{yang2020continuous} addressed trajectory planning for continuous linear systems with discrete control updates, constrained by linear CBF safety sets and STL specifications with linear predicates. In this case, the trajectory planner is based on a Mixed Integer Quadratic Programming (MIQP) formulation that utilizes CBFs to produce system trajectories that are valid in continuous time. In \cite{liang2023control} the authors developed an explicit reference governor-guided CBF (ERG-guided CBF) method that enables the application of first-order CBFs to high-order linearizable systems and enhances STL satisfaction. 

This method reduces the conservativeness of the existing CBF approaches for high-order systems and provides safety guarantees in terms of obstacle avoidance. A further evolution in the application of CBF and STL was presented in \cite{wiltz2022handling}, where the authors circumvented the use of differential inclusions by basing their controller on a set of optimization problems that exploit the piece-wise smoothness of the CBF. This approach reduces the conservativeness of the control method in those points where the CBF is nonsmooth. Consequently, nonsmooth CBFs become applicable to time-varying control tasks, including disjunction operators in their STL fragment. A significant contribution to reducing the computational burden in CBF-based STL motion planning was made in \cite{lindemann2018control} where the control design requires the resolution of a Quadratic Programming (QP) problem during each step of the motion planning process. Furthermore, Lindemann \textit{et al.} extended this CBF-based STL motion planning approach for multi-agent systems \cite{lindemann2019control,lindemann2019decentralized,lindemann2020barrier} with conflicting local specifications and dynamically coupled multi-agent systems.

%contribution
To the best of our knowledge, with reference to the existing literature, the applications in which CBFs are used in conjunction with STL do not allow, due to their construction method, as shown for example in \cite{lindemann2018control}, the completion of a task at any time within a time interval, taking into account the STL ``eventually'' operator. The ``eventually'' operator is a temporal operator that is satisfied if the specification $\phi$ holds at any time before the end of the time interval \cite{bartocci2018specification}. Furthermore, there are no applications where smooth CBF-STL are applied in conjunction with non-linear velocity constraints in safety-critical scenarios, given the presence of both static and dynamic obstacles. In our proposed method, we formulate a smooth CBF-STL control design framework that significantly reduces the conservativeness of existing smooth CBF-STL approaches, potentially allowing the system to operate in a more flexible and efficient manner without compromising safety. 

The contributions of this paper can then be summarized as follows: i) development of a CBF-based STL motion planning methodology for completing a task at any time within a time interval in a dynamic system subject to non-linear velocity constraints, while providing safety-critical guarantees (i.e., velocity constraints and obstacle avoidance); ii) online computation of the smooth CBF-based STL motion planning.

%riassunto next sections
%The rest of the paper is structured as follows: Section \ref{sec:technicalPreliminaries} introduces the required background for the upcoming sections; Section \ref{sec:problemDefinition} presents the problem statement; Section \ref{sec:technicalApproach} describes the proposed approach; Section \ref{sec:simulation} introduces a Matlab simulation in order to validate the proposed CBF-Based Motion Planning approach; Section \ref{sec:conclusion} summarizes the conclusions of this paper.

\section{PRELIMINARIES}
\label{sec:technicalPreliminaries}
In the following, we denote scalars and vectors by non-bold letters $x$ and bold letters $\boldsymbol{x}$, respectively; $\mathbb{R}$ is the set of real numbers, while $\mathbb{R}^n$ is the $n$-dimensional real vector space. Non-negative and positive real numbers are $\mathbb{R}_{\geq0}$ and $\mathbb{R}_{>0}$, respectively. A class $\mathcal{K}$ function $\alpha : \mathbb{R}_{\geq0} \rightarrow \mathbb{R}_{\geq0}$ is a continuous and strictly increasing function with $\alpha(0)=0$. Consider $\boldsymbol{x} \in \mathbb{R}^n$ and $\boldsymbol{u} \in \mathcal{U} \subseteq \mathbb{R}^m$ be the state and input of a non-linear input-affine control system:
\begin{equation} 
    \dot {\boldsymbol {x}}=f(\boldsymbol {x})+g(\boldsymbol {x})\boldsymbol {u}
    \label{eq:modelSystem}
\end{equation}
where the functions $f : \mathbb{R}^n\rightarrow \mathbb{R}^n$ and $g: \mathbb{R}^n\rightarrow \mathbb{R}^{n\times m}$ are locally Lipschitz continuous \cite{lindemann2018control}. Given a control signal $\boldsymbol{u} : [t_0,t_1]\rightarrow \mathcal{U}$, the signal $\boldsymbol{x} : [t_0,t_1]\rightarrow \mathbb{R}^n$ is a solution to (\ref{eq:modelSystem}) if $\boldsymbol{x}$ is absolutely continuous and $\boldsymbol{x}(t)$ satisfies (\ref{eq:modelSystem}) for all $t\in[t_0,t_1]$.
\begin{figure}
        \vspace{5pt}
        \centering
        \includegraphics[width=0.85\linewidth]{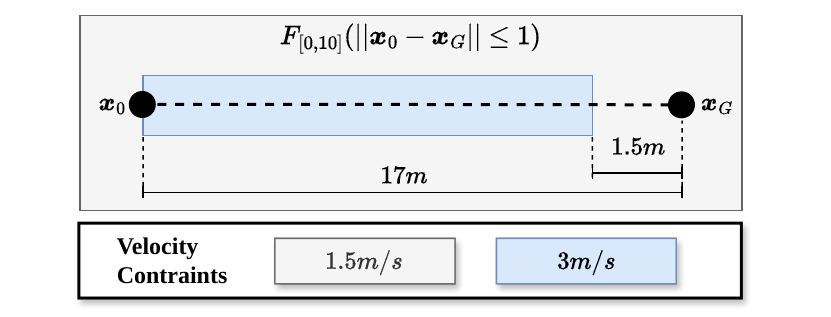}
        \caption{Example 1: The robot must reach the final state $\boldsymbol{x}_G$ within 10 seconds while being subject to two maximum velocity constraints defined by different colored areas along its path.}
        \label{img:example1}
\end{figure}
\begin{figure}
        \centering
        \includegraphics[width=0.85\linewidth]{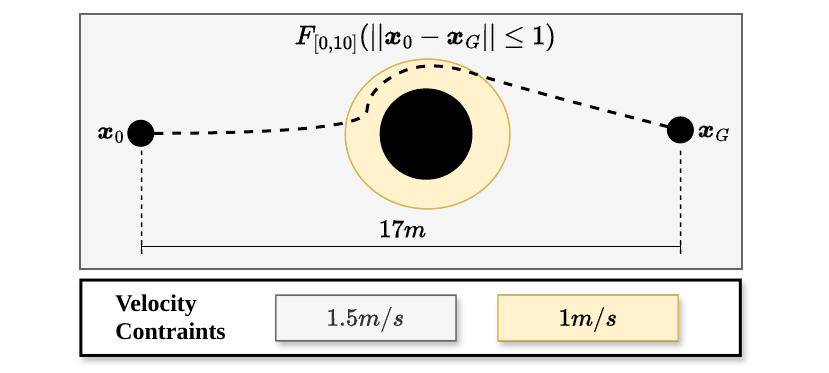}
        \caption{Example 2: The robot must reach the final state $\boldsymbol{x}_G$ within 10 seconds while being subject to two maximum velocity constraints defined by different colored areas along its path in the case of obstacle avoidance.}
        \label{img:example2}
        \vspace{-5pt}
\end{figure}

\subsection{Signal Temporal Logic}
\label{sec:STL}
We utilize STL as a temporal logic formalism due to its capability to express both qualitative and quantitative requirements of systems in continuous domains \cite{zehfroosh2022non,bartocci2018specification}. It offers a natural and compact approach to analyze a robot's motion in a continuously evolving space-time environment. Let $\boldsymbol{s} : \mathbb{R}_{\geq 0} \rightarrow \mathbb{R}^n$ be a continuous-time signal.  STL involves logical predicates, denoted by $\mu$, whose truth values are evaluated over continuous signals $\boldsymbol{s}(t)$. The predicates \cite{maler2004monitoring} are obtained after evaluation of a predicate function $h : \mathbb{R}^n\rightarrow \mathbb{R}$ as:
\begin{equation} 
    \mu \mathrel {\mathrel {\mathop:}\hspace {-0.0672em}=} \begin{cases} \text {True} & \text {if}~ h(\boldsymbol{s}(t))\ge 0\\ \text {False} & \text {if}~ h(\boldsymbol{s}(t))< 0. \end{cases}
\label{eq:predicateSTL}
\end{equation}
In this work, the continuous signal is the system’s state trajectory at time $t$, namely $\boldsymbol{x}(t)$. The STL syntax \cite{lindemann2019control} of an STL formula $\phi$ can be associated with one of the various expressions defined by the grammar in (\ref{eq:syntaxSTL}):
\begin{equation} 
    \phi \mathrel {\mathrel {\mathop:}\!\!{\mathop:}\!\hspace {-0.0672em}=} \top \;|\;\mu \;|\;\neg \phi \;|\;\phi _{1} \wedge \phi _{2}\;|\;\phi _{1} \, U_{[{a},{b}]} \, \phi _{2}.
\label{eq:syntaxSTL}
\end{equation}
In particular, $\phi$ can be associated with the Boolean True ($\top$) signifying that the formula is always true; with the predicate $\mu$ indicating that $\phi$ holds true when the predicate is satisfied as shown in (\ref{eq:predicateSTL}); with the ``negation'' operator meaning that $\phi$ is true when the negated formula $\neg \phi$ is false; with the ``conjunction'' operator of two STL formulas $\phi _{1} \wedge \phi _{2}$ where $\phi$ is true when both $\phi_1$ and $\phi_2$ are simultaneously true; or with the ``until'' temporal operator $\phi _{1} \, U_{[{a},{b}]} \, \phi _{2}$ where $a,b \in \mathbb{R}_{\geq0}$ represent the bounds of the interval defined by the temporal operator, with $a\leq b$. In this context, $\phi$ is true when $\phi_1$ becomes true and remains so within the specified time interval before $\phi_2$ becomes true. To further enhance the expressiveness of STL, two additional operators are introduced: the ``eventually'' temporal operator, defined as $F_{[a,b]}\phi \,:=\, \top \, U_{[{a},{b}]}\, \phi$, and the ``always'' temporal operator, defined as $G_{[a,b]}\phi \,:=\, \neg F_{[a,b]}\neg \phi$, where $G_{[a,b]}\phi$ is satisfied if $\phi$ is not violated during the interval $[a,b]$. The satisfaction relation $(\boldsymbol{x},t)\vDash \phi$ indicates that the signal $\boldsymbol{x} : \mathbb{R}_{\geq 0}\rightarrow \mathbb{R}^n$, e.g., a solution of (\ref{eq:modelSystem}), satisfies $\phi$ at time $t$.

For a signal $\boldsymbol{x} : \mathbb{R}_{\geq0}\rightarrow \mathbb{R}^n$, the STL semantics \cite{lindemann2018control} are defined recursively as follows:
\begin{equation*}
\resizebox{\hsize}{!}{$
\begin{aligned}
&(\boldsymbol {x},t) \models \top \qquad &&\Leftrightarrow \text{holds by definition,} 
\\[0.1cm]
&(\boldsymbol {x},t) \models \mu \qquad &&\Leftrightarrow h(\boldsymbol {x}(t))\ge 0,\\[0.1cm]
&(\boldsymbol {x},t) \models \neg \phi \hspace {-4pt}\qquad &&\Leftrightarrow \neg ((\boldsymbol {x},t) \models \phi),\\
&(\boldsymbol {x},t) \models \phi _{1} \wedge \phi _{2} \hspace {-2.3pt} &&\Leftrightarrow (\boldsymbol {x},t) \models \phi _{1} \wedge (\boldsymbol {x},t) \models \phi _{2},\\[0.1cm]
&(\boldsymbol {x},t) \models \phi _{1} \, U_{[{a},{b}]} \, \phi _{2} &&\Leftrightarrow \exists t_{1} \in [t+a,t+b] \text{ s.t. } (\boldsymbol {x},t_{1})\models \phi _{2},\\[0.1cm]
&&&\quad \; \wedge \forall t_{2}\in [t,t_{1}],(\boldsymbol {x},t_{2}) \models \phi _{1},\\[0.1cm]
&(\boldsymbol {x},t) \models F_{[a,b]}\phi \hspace {1pt} &&\Leftrightarrow \exists t_{1} \in [t+a,t+b] \text{ s.t. } (\boldsymbol {x},t_{1})\models \phi,\\[0.1cm]
&(\boldsymbol {x},t) \models G_{[a,b]}\phi  &&\Leftrightarrow \forall t_{1} \in [t+a,t+b],(\boldsymbol {x},t_{1})\models \phi.
\label{eq:semanticsSTL}
\end{aligned}$}
\end{equation*}
All STL temporal operators have bounded time intervals in continuous time. The horizon of an STL formula is the minimum time needed to decide its satisfaction. For an STL formula that has no nested operators, its horizon is determined by the largest upper bound of the time intervals of all operators  \cite{yang2020continuous}. 

Finally, it is possible to discuss the quality of satisfaction by defining the quantitative semantics $\rho^\phi(\boldsymbol{x},t)\in \mathbb{R}$, which indicates how robustly a signal $\boldsymbol{x}$ satisfies $\phi$ at time $t$ \cite{lindemann2019decentralized}, thus obtaining a robustness value $\rho$ instead of a Boolean value. Furthermore, it holds that $(\boldsymbol{x},t) \vDash \phi$ if $\rho^\phi (\boldsymbol{x},t)>0$ and $(\boldsymbol{x},t)\vDash \phi$ implies $\rho^\phi (\boldsymbol{x},t)\geq0$.

\subsection{Control Barrier Functions encoding STL formulation}
\label{sec:cbf-stl}
A CBF facilitates controller synthesis for dynamic systems by ensuring that, if the system initiates within a specified set, it will always remain within that set. This property establishes the set as forward invariant concerning the system's dynamics. A CBF can define the permissible control inputs that ensure the forward invariance of specific regions for the given dynamical system. 

In \cite{lindemann2020barrier}, the authors have established a connection between a function $\mathfrak {b}: \mathbb{R}^n \times [t_0, t_1] \rightarrow \mathbb{R}$, later shown to be a valid Control Barrier Function (vCBF), and the STL semantics of $\phi$. In particular, if this function is in accordance with the conditions expressed in \cite{lindemann2018control} (i.e., Steps A, B and C in \cite{lindemann2018control}), then, for a given signal $\boldsymbol{x}:\mathbb{R}_{\geq0}\rightarrow \mathbb{R}^n$ with $\mathfrak {b}(\boldsymbol{x}(t),t)\geq0$ for all $t\geq0$, it holds that $(\boldsymbol{x},0)\vDash\phi$. Let the safe set $\mathfrak {C}$ be the set of configurations that satisfy the safety requirements for the system. $\mathfrak {C}$ explicitly depends on time
\begin{align*} 
    \mathfrak {C}(t):=\lbrace \boldsymbol{x}\in \mathbb{R}^n | \mathfrak {b}(\boldsymbol{x},t)\geq 0\rbrace. 
\end{align*}
Hence, $\boldsymbol{x}(t)\in \mathfrak {C}(t)$ for all $t\geq0$ implies $(\boldsymbol{x},0)\vDash\phi$.

In the presence of multiple temporal operators and predicates, we use a smooth approximation of the $\min$ operator. For $p$ functions $\mathfrak {b}_l:\mathbb{R}^n \times \mathbb{R}_{\geq0}\rightarrow \mathbb{R}$ where $l\in\{1,...,p\}$, let $\mathfrak {b}(\boldsymbol{x},t):=-\frac{1}{\eta }\ln \left(\sum _{l=1}^{p} \exp (-\eta \mathfrak {b}_l(\boldsymbol{x},t))\right)$ with $\eta>0$. Note that $\min_{l\in\{1,...,p\}}\mathfrak {b}_l(\boldsymbol{x},t)\approx \mathfrak {b}(\boldsymbol{x},t)$ where the accuracy of this approximation increases as $\eta$ increases, i.e.,
\begin{align*} 
    \lim _{\eta \to \infty }-\frac{1}{\eta }\ln \left(\sum _{l=1}^{p} \exp (-\eta \mathfrak {b}_l(\boldsymbol{x},t))\right)=\min _{l\in \lbrace 1, \ldots ,p\rbrace } \mathfrak {b}_l(\boldsymbol{x},t). 
\end{align*}
Regardless of the choice of $\eta$, we have
\begin{equation} 
    -\frac{1}{\eta }\ln \left(\sum _{l=1}^{p} \exp (-\eta \mathfrak {b}_l(\boldsymbol{x},t))\right)\leq \min _{l\in \lbrace 1, \ldots ,p\rbrace } \mathfrak {b}_l(\boldsymbol{x},t)  
    \label{eq:mixApproximation}
\end{equation}
which is useful since $\mathfrak {b}(\boldsymbol{x},t)\geq0$ implies $\mathfrak {b}_l(\boldsymbol{x},t)\geq0$ for each $l\in\{1,...,p\}$. 

Similar to \cite{lindemann2018control}, a switching mechanism can be used introducing $\mathfrak{o}_l:\mathbb{R}_{\geq0}\rightarrow\{0,1\}$ into $\mathfrak {b}(\boldsymbol{x},t):=-\frac{1}{\eta }\ln \left(\sum _{l=1}^{p} \mathfrak{o}_l(t)\exp (-\eta \mathfrak {b}_l(\boldsymbol{x},t))\right)$; $p$ is again the total number of functions $\mathfrak {b}_l(\boldsymbol{x},t)$ obtained as in \cite{lindemann2018control} (Steps A, B and C) and each $\mathfrak {b}_l(\boldsymbol{x},t)$ corresponds to either an ``always'', ``eventually'', or ``until'' operator with a corresponding time interval $\mathbb{I}=[a_l,b_l]$. If this mechanism is employed, it is necessary to remove the single functions $\mathfrak {b}_l(\boldsymbol{x},t)$ from $\mathfrak {b}(\boldsymbol{x},t)$ when the corresponding ``always'', ``eventually'', or ``until'' operator is satisfied. With these conditions it is possible to synthesize a QP that renders $\mathfrak{C}(t)$ forward invariant when $\mathfrak{b}(\boldsymbol{x},t)$ is a vCBF. Therefore, we can consider: 
\begin{subequations}
    \label{eq:qp_conv1}
    \begin{align}
        &\underset{\hat{\boldsymbol{u}}\in\mathcal{U}}{\operatorname{min}}\; \hat{\boldsymbol{u}}^TQ\hat{\boldsymbol{u}}
        \label{eq:const_qp_cost1}\\
        \begin{split}
            \text{s.t. }
            &\frac{\partial {\mathfrak{b}}({\boldsymbol{x}},t)}{\partial \boldsymbol{x}}(f(\boldsymbol{x})+g(\boldsymbol{x})\hat{\boldsymbol{u}})+\frac{\partial {\mathfrak{b}}({\boldsymbol{x}},t)}{\partial t}\geq-\alpha(\mathfrak{b}({\boldsymbol{x}},t))
            \label{eq:const_qp1}
        \end{split}
    \end{align}
\end{subequations}
where $Q \in \mathbb{R}^{m\times m}$ is a positive semi-definite matrix. This convex optimization problem is feasible if $\mathfrak{b}({\boldsymbol{x}},t)$ is a vCBF. The role of the $\alpha(\cdot)$ function \cite{ferraguti2022safety} is to provide to the designer a way to modulate the action of the CBF, depending on whether a more conservative or aggressive behaviour is desired.
\begin{table}
\vspace{5pt}
    \renewcommand\arraystretch{1.8}
    \centering
    %\caption{Comparison between STL formula and real time employed}
    \caption{Time comparison between constraint in the STL formula and actual path duration. Time in [s].}
    % \caption{Comparison time between STL formula and the real time employed}
    \begin{tabular}{c|c|c|c|c|c}
 	& $\phi'$ & $\phi''$ & $\phi'''$ & $\phi''''$ & $t_{total}$\\
	\hline
	\textbf{STL constraint}  & 10 & 30 & 10 & 10 & 60\\
        \hline
%        \textbf{Real time} & 6.509 s & 25.061 s & 9.01 s &6.499 s & 47.079 s
        \textbf{Actual path duration} &  6.51  &  25.06  & 9.01  &  6.50  &  47.08 
    \end{tabular}
    \label{tab:time_specification}
    \vspace{-5pt}
\end{table}

\section{PROBLEM STATEMENT}
%0.5 pg
\label{sec:problemDefinition}
%definizione problema
We propose a motion planning approach based on CBF and STL for completing a task at any time within a time interval in a dynamic system subject to non-linear velocity constraints. To this end, we consider the STL fragment~\cite{lindemann2020barrier} reported in (\ref{eq:STL-fragment}). In particular, we divide STL formulas into two categories: $\psi$, defining an elementary class, and $\phi$, defining a composite class that includes temporal operators and conjunctions of multiple elementary STL formulas
\begin{subequations}
    \begin{gather}
        \psi::=\top\;|\;\mu\;|\;\neg\mu\;|\;\psi_1 \wedge \psi_2 \label{eq:stl_fragment1}\\
        \phi::=G_{[a,b]}\psi\;|\;F_{[a,b]}\psi\;|\;\psi_1U_{[a,b]}\psi_2\;|\;\phi_1 \wedge \phi_2
        \label{eq:stl_fragment2}
    \end{gather}
    \label{eq:STL-fragment}
\end{subequations}
where $\psi_1$, $\psi_2$ are formulas of the class $\psi$ given in (\ref{eq:stl_fragment1}), whereas $\phi_1$ and $\phi_2$ are formulas of the class $\phi$ given in (\ref{eq:stl_fragment2}). Compared to \cite{lindemann2018control}, we make similar assumptions:
\begin{assumption}
    For an STL formula $\phi$ deﬁned according to (\ref{eq:stl_fragment2}), there exists a constant $C \geq 0$ such that $(\boldsymbol{x}, 0)\vDash \phi \implies ||\boldsymbol{x}(t)|| \leq C \;\forall t \geq 0$.
\end{assumption}
This guarantees that trajectories $\boldsymbol{x}(t)$ are bounded. 
\begin{assumption}
    The vector function $g(\boldsymbol{x})$ in (\ref{eq:modelSystem}) is such that $g(\boldsymbol{x})g(\boldsymbol{x})^T$ is positive deﬁnite for all $\boldsymbol{x} \in \mathbb{R}^n$.
\end{assumption}

Now, the problem under consideration in this paper can be stated as follows.
\begin{problem}
\label{prob:prob1}
   \emph{Given the dynamical system in~\eqref{eq:modelSystem} and an STL formula $\phi$ as in~\eqref{eq:STL-fragment}, derive a control law $\boldsymbol{u}(t)$ so that the solution $\boldsymbol{x}:\mathbb{R}_{\geq 0} \rightarrow \mathbb{R}^n$ to (\ref{eq:modelSystem}) is such that $(\boldsymbol{x},0) \vDash \phi$ providing safety-critical guarantees regarding non-linear velocity constraints and obstacle avoidance.}
\end{problem}

\section{PROPOSED APPROACH}
%1 pg
\label{sec:technicalApproach}
To solve Problem~\ref{prob:prob1}, we compute a valid CBF $\mathfrak {b}(\boldsymbol{x},t)$ at any time and leverage the ``eventually'' operator. For this reason in Section \ref{subsec:onlineb} we explain the general procedure for how to dynamically calculate the vCBF when $\phi := F_{[t_a,t_b]}\mu_l$ or $\phi := G_{[t_a,t_b]}\mu_l$ where $\mu_l$ does not contain any conjunctions, i.e., considering the definition given in~\eqref{eq:mixApproximation}, \mbox{$  p=1$}. Subsequently, in Section \ref{subsec:metodology} we present the proposed CBF-based STL motion planning methodology.

 % \subsection{Online computation of the smooth CBF-based STL motion planning}
\subsection{Online computation of smooth CBF-STL motion planning}
\label{subsec:onlineb}
%Calcolo online b

The construction of $\mathfrak {b}(\boldsymbol{x},t)$ is equivalent to the procedure used in \cite{lindemann2020barrier}; we refer interested readers to the original paper for further details. For completeness, the procedure is summarized as follows:
\begin{itemize}
    \item Consider $\phi := F_{[t_a,t_b]}\mu$ or $\phi := G_{[t_a,t_b]}\mu$ and let
    \begin{equation}
        t^* :=\bigg \{
        \begin{array}{rl}
        t_b & \text{if} \; F_{[t_a,t_b]}\mu \; , \;t_b > 0\\
        t_a & \text{if} \; G_{[t_a,t_b]}\mu \; , \;t_a \geq 0.
        \end{array}
        \label{eq:t*}
    \end{equation}
    \item Let
    \begin{equation}
        h^{opt}:=\sup_{\boldsymbol{x}\in\mathbb{R}^n}h(\boldsymbol{x}).
        \label{hopt}
    \end{equation}
    \item Since we aim at satisfying $\phi$ with robustness threshold $r \in \mathbb{R}_{\geq 0}$, i.e., $\rho^\phi (\boldsymbol{x},0) \geq r$, then choose
    \begin{equation}
        r \in \bigg \{
        \begin{array}{rl}
        (0,h^{opt}) & \text{if} \; t^* > 0\\
        (0,h(\boldsymbol{x}(0))) & \text{if} \; t^* \geq 0.
        \end{array}
        \label{eq:r}
    \end{equation}
    \item Consider $\mathfrak {b}(\boldsymbol{x},t) := -\gamma(t) + h(\boldsymbol{x})$ where $\gamma(t)$ is a non-decreasing function defined as piecewise linear function
    \begin{equation}
        \gamma(t) := \bigg \{
        \begin{array}{rl}
        \frac{\gamma_\infty - \gamma_0}{t^*}t+\gamma_0 & \text{if} \; t<t^*\\
        \gamma_\infty & \text{otherwise}.
        \end{array}
        \label{eq:gamma}
    \end{equation}

    \item Next, let
    \begin{subequations}
    \begin{gather}
        \gamma_0 \in (-\infty,h(\boldsymbol{x}(0))
        \label{eq:gamma0},\\
        \gamma_\infty \in (\max(r,\gamma_0),h^{opt}).
        \label{eq:gamma00}
    \end{gather}
    \label{eq:gamma0-00}
    \end{subequations}
\end{itemize}
By means of this procedure, it is possible to obtain $\mathfrak{b}({\boldsymbol{x}},t)$, and consequently $\frac{\partial {\mathfrak{b}}({\boldsymbol{x}},t)}{\partial \boldsymbol{x}}$ and $\frac{\partial {\mathfrak{b}}({\boldsymbol{x}},t)}{\partial t}$ in order to achieve the function in (\ref{eq:const_qp1}).

\begin{figure}[tb]
\vspace{5pt}
      \centering
      \includegraphics[width=0.47\textwidth]{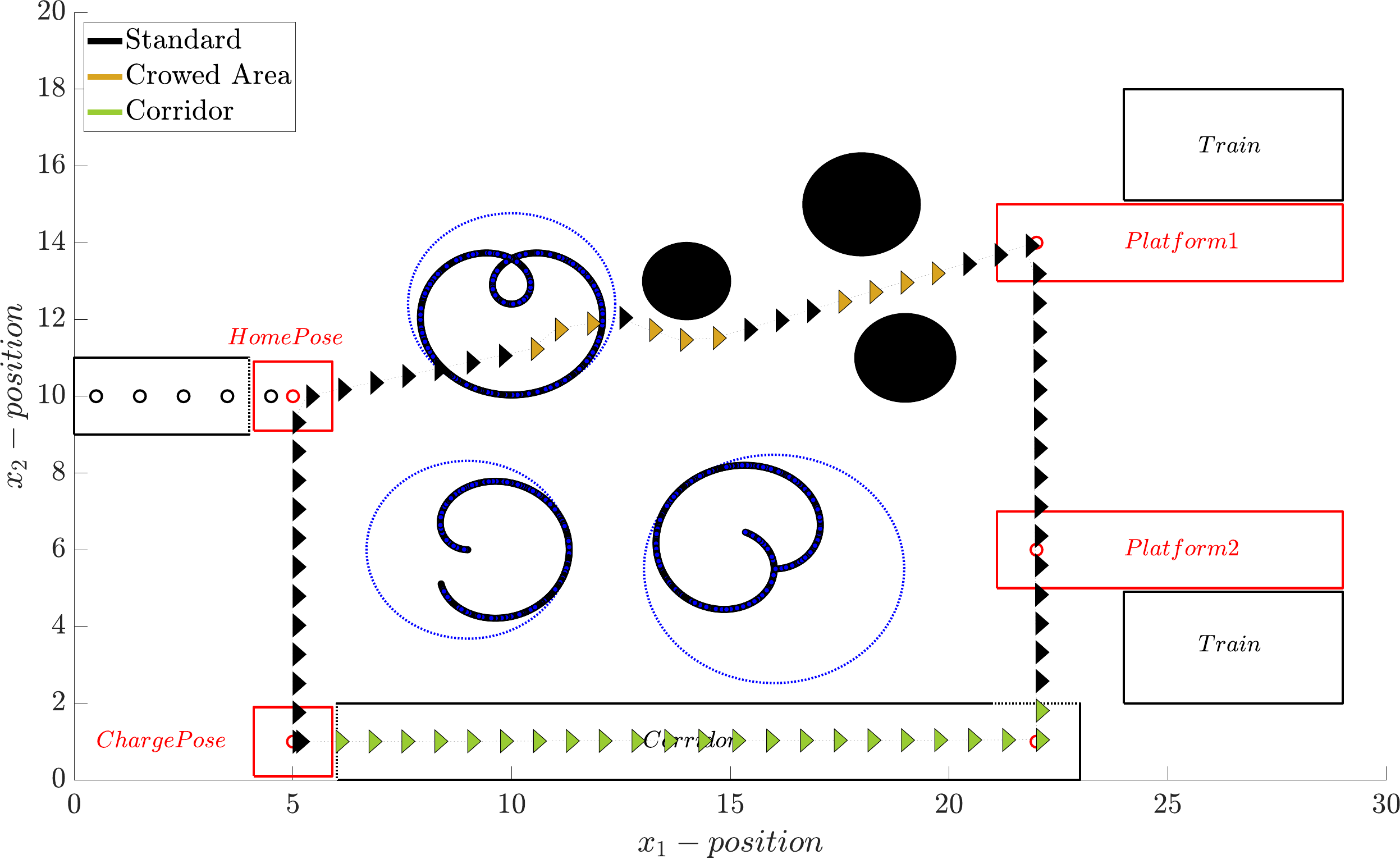}
      \caption{Robot trajectory in simulated environment.}
      \label{img:simulation}
\end{figure}
\begin{figure}[tb]
      \centering
      \includegraphics[width=0.47\textwidth]{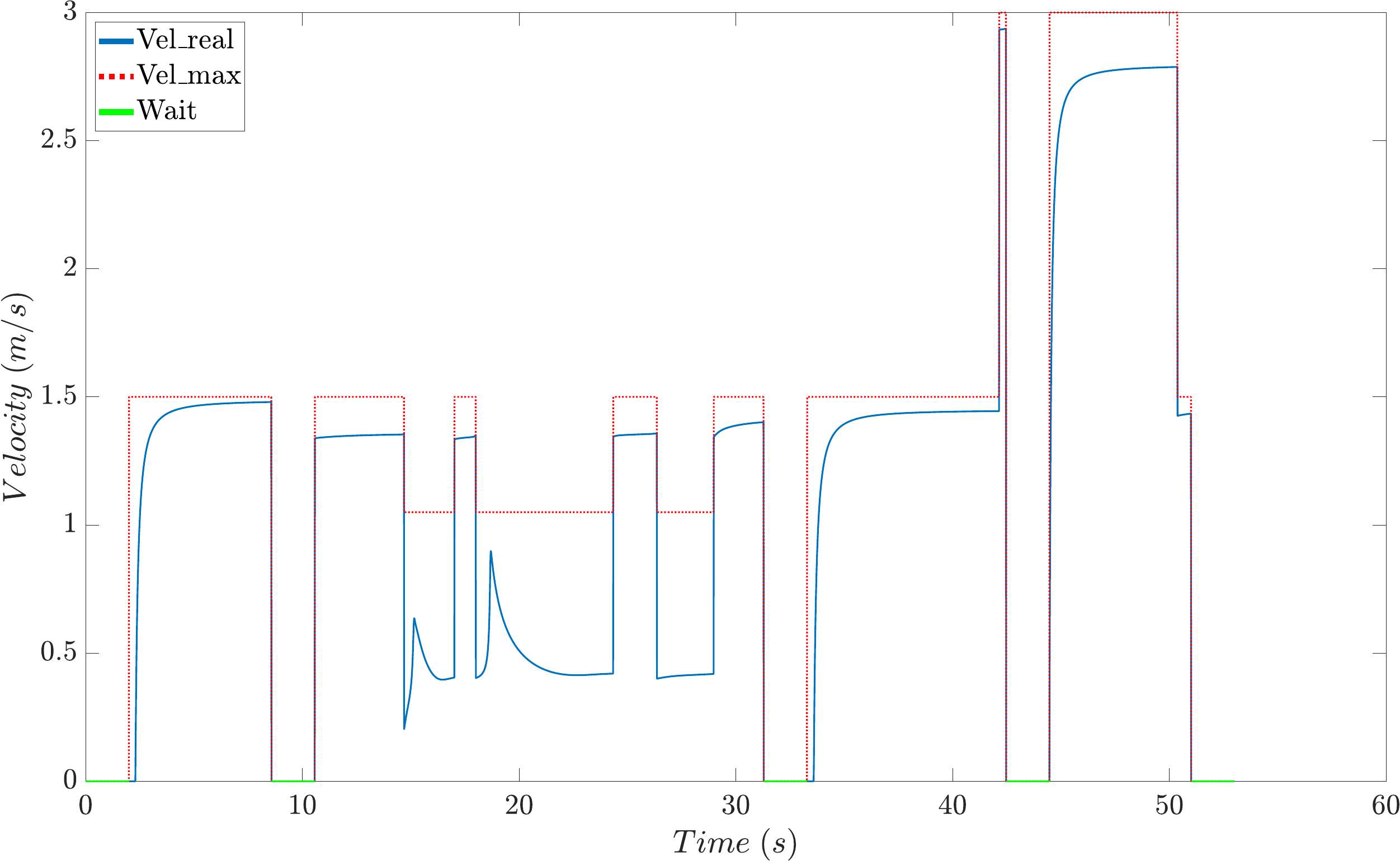}
      \caption{Time evolution of real speed in reference to maximum speed.}
      \label{img:velocity}
      \vspace{-5pt}
\end{figure}
\begin{figure}[tb]
\vspace{5pt}
      \centering
      \includegraphics[width=0.46\textwidth]{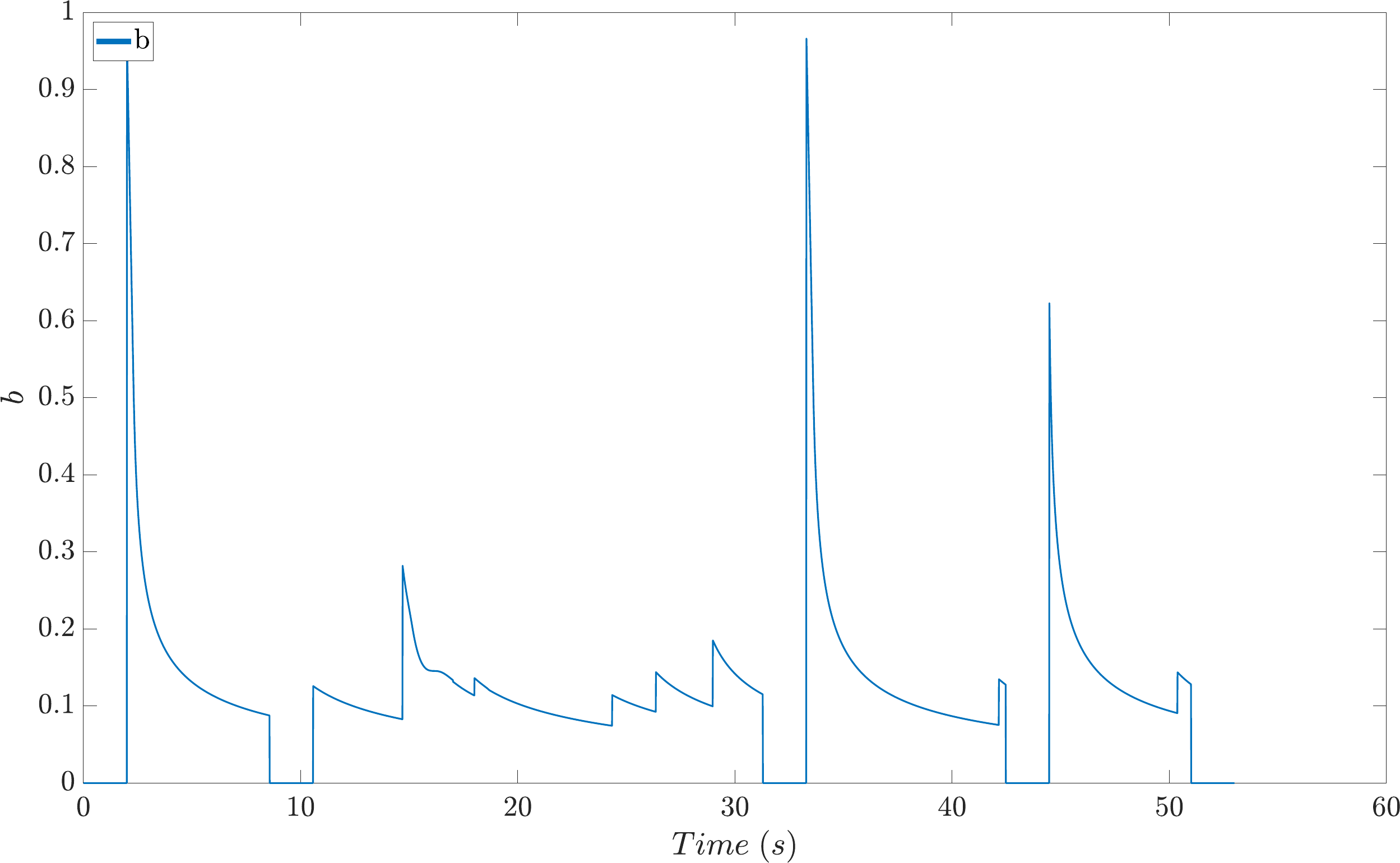}
      \caption{The behavior of the control barrier function $\mathfrak{b}({\boldsymbol{x}},t)$ has a value always greater than zero, indicating the satisfaction of the STL formula $\phi$.}
      \label{img:b}
      \vspace{-5pt}
\end{figure}

\subsection{CBF-based STL motion planning methodology}
\label{subsec:metodology}
For a dynamical system, safety constraints can be thought of as those delimiting a safe region of its state space, in which the state must remain all the time. Such a region is what we define here as the safe set \cite{ferraguti2022safety}. Let us consider having a robot in the initial position $\boldsymbol{x}_0=[x_{0_x},x_{0_y}]^T$ and it needs to reach the goal $\boldsymbol{x}_G=[x_{G_x},x_{G_y}]^T$ in a time $t^*=t_b$ seconds with a certain tolerance distance $\varepsilon$ from $\boldsymbol{x}_G$. This problem can be expressed in terms of the following STL formula:
\begin{equation*}
    \phi : F_{[t_a,t_b]}(||\boldsymbol{x}_0-\boldsymbol{x}_G||<\varepsilon).
\end{equation*}
To construct $\mathfrak {b}(\boldsymbol{x},t)$, in the case of the ``eventually'' operator, according to Sec.~\ref{subsec:onlineb}, it is necessary to consider $t^*$ as the time $t_b$ belonging to the interval $\mathbb {I}=[t_a;t_b]$. This process does not allow the desired task to be completed at any time within the interval $\mathbb {I}$, but it will be completed at $t_b$, unless the vCBF is constructed with a different $t^*<t_b \in \mathbb {I}$. 
Furthermore, considering a system that must provide safety-critical guarantees, it is necessary to consider the possibility that it may be subject to velocity constraints and that it must complete the task while ensuring collision avoidance and remaining within the safe set. We propose the following CBF-based STL motion planning methodology that allows solving these problems. It consists of two construction steps.

\subsubsection{STEP 1} 
The first step involves determining the average velocity, $v_{average}$, that the robot would have along the path to be crossed. Assuming there are no obstacles, the robot will follow the minimum Euclidean distance path, thus obtaining that $v_{average}$, set by the STL formula $\phi$, is given by the ratio between the variation of the distance traveled $\Delta s_{tot}$, and the time $t^*$ imposed by $\phi$, as defined in (\ref{eq:STEP1}):
\begin{subequations}
    \begin{gather}
        \Delta s_{tot} = \sqrt{(x_{0_x}-x_{G_x})^2+(x_{0_y}-x_{G_y})^2}
        \label{eq:space},\\
        v_{average}=\frac{\Delta s_{tot}}{t^*}.
        \label{vm}
    \end{gather}
    \label{eq:STEP1}
\end{subequations}
In the event that one or more obstacles are present along the robot's path, the distance it will need to traverse will be greater, leading to an increase in velocity to ensure the satisfaction of the STL specification $\phi$ (i.e., to ensure that the task is completed by $t^*$).
Introducing the velocity constraint $v_{max}$, two sub-cases can be considered: the first, $v_{max} \geq v_{average}$ in which the specification $\phi$ continues to be satisfied; the second, $v_{max} < v_{average}$, in which the STL specification $\phi$ cannot be satisfied. Consider, for example, the case shown in Fig. \ref{img:example1}: in this situation, the STL specification requires the robot to reach the destination $\boldsymbol{x}_G$ from the initial position $\boldsymbol{x}_0$, covering a distance of $\Delta s_{tot} = 17 \; m$, in $t^* = 10 \;s$. As a result, $v_{average}=1.7 \;m/s$, which does not allow the satisfaction of the specification in the last tract of $1.5 \;m$, shown in grey, since $v_{max}<v_{average}$. A similar scenario that considers the presence of an obstacle is depicted in Fig.~\ref{img:example2}. In this case as well, the robot should have an average velocity $v_{average}=1.7 \;m/s$ to ensure the satisfaction of the specification within $t^* = 10 \;s$. This cannot be achieved as it conflicts with the velocity constraint of $v_{max} = 1\; m/s$ when the robot enters in the area shown in yellow near the obstacle, since $v_{max}<v_{average}$.

\subsubsection{STEP 2}
This step provides a solution that allows resolving the issue introduced by the velocity constraint and simultaneously addresses the problem associated with the inability to conclude a task at any time within the interval $\mathbb{I}$, considering the ``eventually'' operator. 

In particular, we propose the computation of a dynamically defined bound $t^*_{new}$, given by
\begin{equation}
     t_{new}^*=\frac{\Delta s}{(P_i - P_r P_c)\;v_{max}(t)}
    \label{eq:STEP2}
\end{equation}
where:
\begin{itemize}
\item $\Delta s$ is the remaining path space;
\item $v_{max}(t)$ is the maximum velocity;
\item $P_i$ represents the \emph{initial percentage}, taking a user-defined value $0.5 \leq P_i < 1$ which allows the $v_{average}$ to be increased so as to complete the specification within $\mathbb{I}$. In the simulation presented in Section \ref{sec:simulation}, $P_i$ is set to 0.9;
\item $P_r$ is the \emph{reduction percentage}, taking an arbitrary value $0 < P_i < 0.2$ which allows to decrement \emph{initial percentage} in case the solver does not converge to solution. In the simulation in Section \ref{sec:simulation}, $P_i$ is considered as 0.025;
\item $P_c$ is the \emph{percentage counter}, with an initial value of zero, and it is incremented in case the QP fails to find a solution.
\end{itemize}

Specifically, the solution consists in instantaneously checking the maximum velocity $v_{max}(t)$ and dynamically computing a $t^*_{new}$ based on the remaining distance $\Delta s$ and the weighted maximum allowable velocity $(P_i - P_r P_c)\;v_{max}(t)$.
This computation needs to be implemented whenever a change in $v_{max}$ occurs or when the quadratic problem does not converge to a feasible solution due to the exceeding of speed constraints. 
Hence, the quantity $t^*_{new}$ will be used to compute a new barrier $\mathfrak{b}({\boldsymbol{x}},t)$, as explained in Section \ref{subsec:onlineb}. This procedure will allow the robot to travel the route at a speed greater than $v_{average}$, thus allowing the specification to be satisfied in the interval $\mathbb{I}$, and to overcome the problem depicted in Figs.~\ref{img:example1} and~\ref{img:example2}. 

Regarding non-linear velocity constraints, considering the velocity vector $\boldsymbol{v}_{real} = [v_x,v_y,w]^T$, we introduce an additional constraint within the QP problem by imposing the squared norm of the non-linear velocity to be \mbox{$\left \|v_x + v_y\right \| \leq v_{max}$}.
Therefore, the overall SRN problem can be formulated as follows
\begin{subequations}
    \label{eq:qp_conv2}
    \begin{align}
        &\underset{\hat{\boldsymbol{u}}\in\mathcal{U}}{\operatorname{min}}\; \hat{\boldsymbol{u}}^TQ\hat{\boldsymbol{u}}
        \label{eq:const_qp_cost2}\\
        \begin{split}
            \text{s.t. }
            &\frac{\partial {\mathfrak{b}}({\boldsymbol{x}},t)}{\partial \boldsymbol{x}}(f(\boldsymbol{x})+g(\boldsymbol{x})\hat{\boldsymbol{u}})+\frac{\partial {\mathfrak{b}}({\boldsymbol{x}},t)}{\partial t}\geq-\alpha(\mathfrak{b}({\boldsymbol{x}},t))\\
            &\left \|v_x + v_y\right \| \leq v_{max}.
            \label{eq:const_qp2}
        \end{split}
    \end{align}
\end{subequations}
\section{SIMULATION RESULTS}
\label{sec:simulation}
In order to test the proposed framework, we simulated an SRN application in Matlab, shown in Fig.~\ref{img:simulation} and in the accompanying video\footnote{\href{https://doi.org/10.5281/zenodo.10075373}{DOI: https://doi.org/10.5281/zenodo.10075373}}.
Let us consider a scenario in which a robot is located in a train station and needs to guide a person to a platform. Depending on the robot's position, the velocity constraint $v_{max}$ can vary, creating different operational modes:
\begin{itemize}
    \item \emph{Standard}: the default mode, with $v_{max}= 1.5 \;m/s$;
    \item \emph{Crowded Area}: when the robot is close to dynamic or static obstacles, with $v_{max}= 1.05 \;m/s$;
    \item \emph{Corridor}: when the robot is within a corridor, where human access is not allowed and we assume there are not obstacles. Here, we set $v_{max}= 3 \;m/s$.
\end{itemize}

For the simulation, we employed a three-wheeled omnidirectional robot model as implemented in \cite{lindemann2018control}. The simulation environment incorporates six obstacles, comprising three static and three dynamic obstacles, randomly distributed in the environment. To create a more realistic simulation of dynamic obstacle behavior, distinct rhodonea curve trajectories are assigned to each obstacle. To successfully guide the user to their destination, the robot starts from the \emph{charge\_pose} and proceeds to the \emph{home\_pose}. Subsequently, it must safely guide the user to the platform while avoiding collisions. Upon completing the task, the robot will enter the \emph{corridor} and ultimately return to the initial position. The robot is expected to execute this sequence of operations within various time intervals, for a maximum total of 60 seconds. The temporal constraints are expressed through the following STL formula: $\phi=\phi'\wedge \phi''\wedge \phi'''\wedge \phi''''$ with a certain tolerance $\varepsilon=0.2 \;m$ and velocity constrains, where:
\begin{equation*}
    \begin{aligned}
        \phi' :&= F_{[0,10]}(||\boldsymbol{x}-\boldsymbol{x}_{HOME\_POSE}||\leq \varepsilon)
        \\
        \phi'' :&= F_{[10,40]}(||\boldsymbol{x}-\boldsymbol{x}_{PLATFORM\_1}||\leq \varepsilon)
        \\
        \phi''' :&= F_{[40,50]}(||\boldsymbol{x}-\boldsymbol{x}_{CORRIDOR}||\leq \varepsilon)
        \\
        \phi'''' :&= F_{[50,60]}(||\boldsymbol{x}-\boldsymbol{x}_{CHARGE\_POSE}||\leq \varepsilon).
    \end{aligned}
\end{equation*}
Figure~\ref{img:velocity} shows the real velocity profile along the path that satisfies the non-linear velocity constraints, shown in red. For simplicity, we considered the robot internal dynamics $f(\boldsymbol{x)}$ equal to zero. As a result, its velocity  is given by $\boldsymbol{v}_{real}=g(\boldsymbol{x})\boldsymbol{u}$, from~\eqref{eq:modelSystem}.
During the simulation, pauses were introduced at the end of each STL specification, as indicated by the green segments in Fig.~\ref{img:velocity}, where the robot is not moving. In addition, as reported in Table~\ref{tab:time_specification}, it can be observed that each specification is satisfied within the time interval defined by its own temporal operator. As a result, by employing this approach, it was possible to decrease the execution time of the STL formula $\phi$ to approximately $47\;s$ instead of $60\;s$. The performance of the function $\mathfrak{b}({\boldsymbol{x}},t)$, shown in Fig. \ref{img:b}, demonstrates the satisfaction of the formula $\phi$ throughout the entire simulation, as it ensures that its value is always greater than zero. 
Using the proposed approach, it is possible to observe the results of motion planning, which has allowed for the identification of a valid path for the robot and the satisfaction of STL specifications subject to non-linear velocity constraints, ensuring the compliance with safety guarantees.

\section{CONCLUSION}
\label{sec:conclusion}
We proposed a CBF-based STL motion planning methodology for completing a task at any time within a time interval $\mathbb{I}$ in a dynamic system subject to non-linear velocity constraints, while providing safety-critical guarantees given the presence of both static and dynamic obstacles. 
This procedure uses an online computation of the smooth CBF using a value, dynamically calculated, $t_{new}^*$ based on the remaining path space and the weighted maximum allowable velocity. Next, a simulation was proposed in order to validate the methodology showing proper compliance with spatio-temporal constraints subject to non-linear velocity constraints. This research opens new possibilities for safety-critical navigation in complex environments by leveraging the combination of STL and CBFs in a computationally-efficient manner. 
In the future, we plan to incorporate additional types of constraints for use within a social-navigation context, such as the constraint on the robot's rotation to ensure that the user is always within the robot's field of view. Furthermore, this architecture will be implemented in a mobile robot for experimental validation in physical environment.




% \section*{ACKNOWLEDGMENT}

% ...

%%%%%%%%%%%%%%%%%%%%%%%%%%%%%%%%%%%%%%%%%%%%%%%%%%%%%%%%%%%%%%%%%%%%%%%%%%%%%%%%

\bibliographystyle{unsrt}
\bibliography{main}

\begin{thebibliography}{10}

\bibitem{LidarSlam}
K.~Song, Y.~Chiu, L.~Kang, S.~Song, C.~Yang, P.~Lu, and S.~Ou.
\newblock Navigation control design of a mobile robot by integrating obstacle avoidance and lidar slam.
\newblock In {\em Int. Conf. on Syst., Man, and Cybern.} IEEE, 2018.

\bibitem{onlineRobotNavigation}
S.~Silva, N.~Verdezoto, D.~Paillacho, S.~Millan-Norman, and J.~D. Hern{\'a}ndez.
\newblock Online social robot navigation in indoor, large and crowded environments.
\newblock In {\em Int. Conf. on Robot. and Autom.} IEEE, 2023.

\bibitem{maler2004monitoring}
O.~Maler and D.~Nickovic.
\newblock Monitoring temporal properties of continuous signals.
\newblock In {\em Int. Symp. on Formal Techniques in Real-Time and Fault-Tolerant Syst.} Springer, 2004.

\bibitem{wiltz2022handling}
A.~Wiltz and D.~V Dimarogonas.
\newblock Handling disjunctions in signal temporal logic based control through nonsmooth barrier functions.
\newblock In {\em Conf. on Decision and Control}. IEEE, 2022.

\bibitem{belta2007symbolic}
C.~Belta, A.~Bicchi, M.~Egerstedt, E.~Frazzoli, E.~Klavins, and G.~J Pappas.
\newblock Symbolic planning and control of robot motion [grand challenges of robotics].
\newblock {\em Robot. \& Autom. Magazine}, 2007.

\bibitem{9147796}
L.~Lindemann and D.~V. Dimarogonas.
\newblock Efficient automata-based planning and control under spatio-temporal logic specifications.
\newblock In {\em Am. Control Conf.}, 2020.

\bibitem{liang2023control}
K.~Liang, M.~Cai, and C.~Vasile.
\newblock Control barrier function for linearizable system with high relative degrees from signal temporal logics: A reference governor approach.
\newblock {\em arXiv preprint arXiv:2309.08813}, 2023.

\bibitem{wieland2007constructive}
P.~Wieland and F.~Allg{\"o}wer.
\newblock Constructive safety using control barrier functions.
\newblock {\em IFAC Proceedings Volumes}, 2007.

\bibitem{nguyen2016exponential}
Q.~Nguyen and K.~Sreenath.
\newblock Exponential control barrier functions for enforcing high relative-degree safety-critical constraints.
\newblock In {\em Am. Control Conf.} IEEE, 2016.

\bibitem{zehfroosh2022non}
A.~Zehfroosh and H.~G Tanner.
\newblock Non-smooth control barrier navigation functions for stl motion planning.
\newblock {\em Frontiers in Robot. and AI}, 2022.

\bibitem{yang2020continuous}
G.~Yang, C.~Belta, and R.~Tron.
\newblock Continuous-time signal temporal logic planning with control barrier functions.
\newblock In {\em Am. Control Conf.} IEEE, 2020.

\bibitem{lindemann2018control}
L.~Lindemann and D.~V Dimarogonas.
\newblock Control barrier functions for signal temporal logic tasks.
\newblock {\em Control Syst. letters}, 2018.

\bibitem{lindemann2019control}
L.~Lindemann and D.~V Dimarogonas.
\newblock Control barrier functions for multi-agent systems under conflicting local signal temporal logic tasks.
\newblock {\em Control Syst. letters}, 2019.

\bibitem{lindemann2019decentralized}
L.~Lindemann and D.~V Dimarogonas.
\newblock Decentralized control barrier functions for coupled multi-agent systems under signal temporal logic tasks.
\newblock In {\em Eur. Control Conf.} IEEE, 2019.

\bibitem{lindemann2020barrier}
L.~Lindemann and D.~V Dimarogonas.
\newblock Barrier function based collaborative control of multiple robots under signal temporal logic tasks.
\newblock {\em Transactions on Control of Netw. Syst.}, 2020.

\bibitem{bartocci2018specification}
E.~Bartocci, J.~Deshmukh, A.~Donz{\'e}, G.~Fainekos, O.~Maler, D.~Ni{\v{c}}kovi{\'c}, and S.~Sankaranarayanan.
\newblock Specification-based monitoring of cyber-physical systems: a survey on theory, tools and applications.
\newblock {\em Lectures on Runtime Verification: Introductory and Advanced Topics}, 2018.

\bibitem{ferraguti2022safety}
F.~Ferraguti, C.~T. Landi, A.~Singletary, H.~Lin, A.~Ames, C.~Secchi, and M.~Bonf{\`e}.
\newblock Safety and efficiency in robotics: the control barrier functions approach.
\newblock {\em Robot. \& Autom. Magazine}, 2022.

\end{thebibliography}

\end{document}